%% file: main.tex
\documentclass[10pt,twocolumn,letterpaper]{article}

\usepackage{cvpr}              
\input{preamble}

\definecolor{cvprblue}{rgb}{0.21,0.49,0.74}
\usepackage[pagebackref,breaklinks,colorlinks,allcolors=cvprblue]{hyperref}

\usepackage{multirow}
\usepackage{float}

\usepackage{pifont}
\newcommand{\cmark}{\textcolor{green}{\ding{51}}}
\newcommand{\xmark}{\textcolor{red}{\ding{55}}}

\usepackage{graphicx}

\usepackage{background}

\newcommand{\mywatermark}{%
    \begin{minipage}{\textwidth}
        \centering
        \fontsize{10}{10}\selectfont % Adjust font size and baselineskip
        This paper has been accepted at the 5th International Workshop on Event-Based Vision (CVPR 2025) 
    \end{minipage}%
}

\backgroundsetup{
    scale=1,
    color=gray,
    angle=0,
    opacity=0.5,
    position=current page.north,
    vshift=-1cm, % Adjust vertical position
    contents={\mywatermark}
}

\def\paperID{46} % *** Enter the Paper ID here
\def\confName{CVPR}
\def\confYear{2025}

\title{Event Quality Score (EQS): Assessing the Realism of Simulated Event Camera Streams via Distances in Latent Space}

\author{Kaustav Chanda, Aayush Atul Verma, Arpitsinh Vaghela, Yezhou Yang, Bharatesh Chakravarthi \\
Arizona State University \\
{\tt\small \{kchanda3, averma90, avaghel3, yz.yang, bshettah\}@asu.edu}\
}

\begin{document}

\twocolumn[{%
\renewcommand\twocolumn[1][]{#1}%
\maketitle
\begin{center}
    \centering    
    % \fbox{\rule{0pt}{2in} \rule{.92\linewidth}{0pt}} 
    \includegraphics[width=0.95\linewidth]{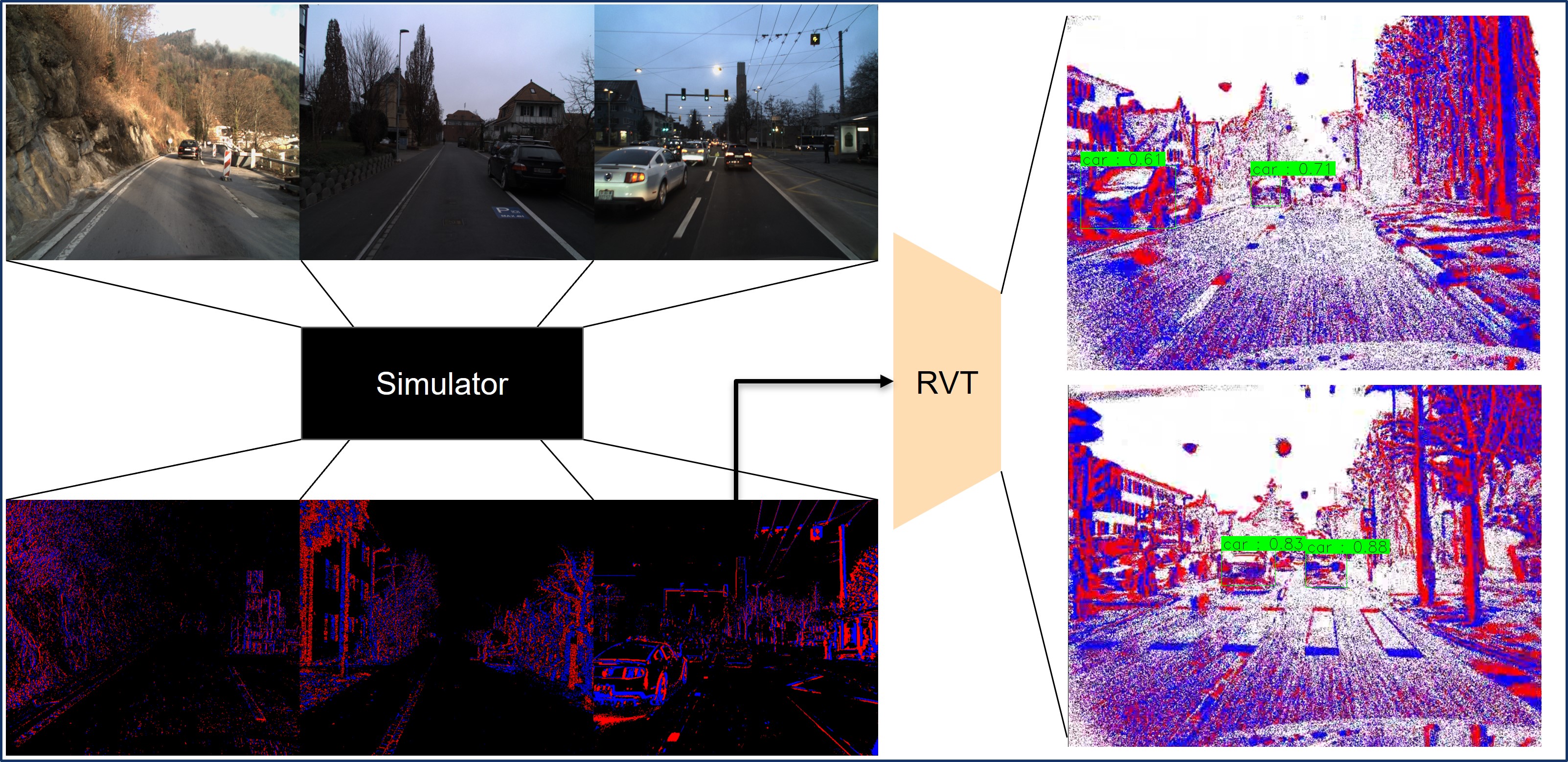}
    \captionof{figure}{Schematic representation of a recurrent vision transformer (RVT) model trained on simulated event data and evaluated on real event streams.}
    \label{fig_teaser}
% % \end{figure}
\end{center}%
}]

% \maketitle
\input{sec/0_abstract}    
\input{sec/1_intro}

\input{sec/2_related}

\input{sec/3_EQS}
\input{sec/4_Experiments}

\input{sec/5_Conclusion}

{
    \small
    \bibliographystyle{ieeenat_fullname}
    \bibliography{main}
}

% WARNING: do not forget to delete the supplementary pages from your submission 
% \input{sec/X_suppl}

\end{document}

%% file: preamble.tex
\usepackage{lineno}

%% file: sec/0_abstract.tex
\begin{abstract}

Event cameras promise a paradigm shift in vision sensing with their low latency, high dynamic range, and asynchronous nature of events. Unfortunately, the scarcity of high-quality labeled datasets hinders their widespread adoption in deep learning-driven computer vision. To mitigate this, several simulators have been proposed to generate synthetic event data for training models for detection and estimation tasks. However, the fundamentally different sensor design of event cameras compared to traditional frame-based cameras poses a challenge for accurate simulation. As a result, most simulated data fail to mimic data captured by real event cameras. Inspired by existing work on using deep features for image comparison,  we introduce \textit{event quality score} ($EQS$), a quality metric that utilizes activations of the RVT architecture. Through sim-to-real experiments on the DSEC driving dataset, it is shown that a higher $EQS$  implies improved generalization to real-world data after training on simulated events. Thus, optimizing for $EQS$ can lead to developing more realistic event camera simulators, effectively reducing the simulation gap. EQS is available at \href{https://github.com/eventbasedvision/EQS}{https://github.com/eventbasedvision/EQS} .

\end{abstract}

%% file: sec/1_intro.tex
\section{Introduction}
\label{sec:intro}

Event cameras \cite{chakravarthi2024recent} are neuromorphic sensors inspired by the human retina, where rods and cones activate independently based on the light stimulus received in real-time. Analogous to this mode of operation, event cameras asynchronously register pixel-level changes in the scene with precise timestamps in the order of microseconds. This results in a spatially sparse but temporally dense data stream. Additionally, the absence of a fixed exposure time frame makes these sensors resilient to drastic changes in lighting conditions with a dynamic range in the region of $120$ $dB$ \cite{delbruck_davis}. These characteristics make them ideal for vision applications in challenging tasks such as object detection in low-light \cite{liu2024seeingmotionnighttimeevent,li2024eventassistedlowlightvideoobject, chakravarthi2023event}, over-exposure \cite{gehrig2021dsecstereoeventcamera}, high-speed motor control \cite{muglikar2024eventcamerasmeetspads, Liu_2023}, and autonomous landing and flight \cite{Pijnacker_Hordijk_2017, Vidal_2018}.

% characteristics
A single pixel of an event camera consists of three components \cite{delbruck_davis}: a photo receptor circuit, a differencing circuit, and a comparator. The photo-receptor circuit converts photocurrent logarithmically into voltage, and the difference between these values is compared against global thresholds in the comparator to determine when to fire an event. Through this process, shot noise from photons, thermal noise from transistors, and other non-idealities in clocking and quantization \cite{gallego_survey} impart a stochastic nature to the data captured by event cameras that exacerbates the challenge for models trained on simulated data to generalize.  

One of the key enabling factors for the deep learning revolution in computer vision has been the availability of large-scale image datasets captured using widely available sensors in cameras and smartphones. Due to their limited adoption across the industry, there is an acute scarcity of large, high-quality datasets captured using these sensors. To mitigate this limitation, several event camera simulators have been proposed 
\cite{hu2021v2evideoframesrealistic,pix2nvs,Mueggler2017,han2024physical,dosovitskiy2017carlaopenurbandriving, ICNS, Lin2022}. These simulators have been utilized both for generating new datasets such as SEVD \cite{aliminati2024sevd} as well as event-based counterparts of existing image datasets such as Event-KITTI \cite{event_kitti}, Yelan$-$Syn \cite{yelan_syn_dataset}, ESfP$-$Synthetic \cite{Muglikar23CVPR_esfp}, N$-$EPIC Kitchen and  N$-$ImageNet \cite{kim2022nimagenetrobustfinegrainedobject}. 
Although these datasets have motivated much of the initial research in event vision, the efficiency of models trained using synthetic datasets on real-world data has not been thoroughly investigated. 

% why diff - existing approaches - why wrong
The primary difficulty in characterizing event streams arises from how event data is structured. An event stream is an array of $\left<x, y, p, t\right>$ tuples, each indicating that the log intensity at the pixel location $(x,y)$ during the timestamp $t$ changed by a factor greater than $pol \times C$ where $C$ is a constant threshold for a given scene. Therefore, existing approaches convert event streams into an image-like $2D$ representation before applying well-known image similarity metrics such as structural similarity (SSIM) \cite{ssim} or peak signal-to-noise ratio (PSNR). SSIM and PSNR have been extensively studied, and it has been shown in \cite{Stoffregen2020} that not only are they unsuitable for directly comparing event frames but also produce misleading results for grayscale frames reconstructed from event streams with encoder-decoder models. Crucially, no measurement framework exists that computes a differentiable similarity score directly between two event streams without relying on an intermediate representation.

The LPIPS \cite{lpips} metric proposed for image comparison has been shown to alleviate some of the aforementioned shortcomings of traditional similarity metrics. The authors show that the internal activations of networks trained for high-level tasks such as classification are capable of capturing the subtle differences in input data, such as blurriness, and that these activations correlate strongly with human perception. \cite{rebecq2019eventstovideobringingmoderncomputer} utilized the LPIPS loss for training a UNet \cite{ronneberger2015unetconvolutionalnetworksbiomedical} architecture for reconstructing grayscale images from event frames, indicating that the activations of a VGG \cite{simonyan2015deepconvolutionalnetworkslargescale_vgg} model can serve as an effective loss function. \cite{Stoffregen2020} showed that the noise model used in simulated event data has a significant impact on model performance and that a simple zero-mean Gaussian noise to simulate background events and a few uniformly selected `hot' pixels improved model performance for grayscale reconstruction from simulated events. However, these approaches focus solely on the accuracy of reconstructed image frames. Our framework directly evaluates the quality of event streams analogous to the approach used by \cite{lpips} for frames without converting to a $2D$ image-like representation first.

The proposed \textit{event quality score ($EQS$)} measure takes two event streams of arbitrary length as input and converts them to a $4$-dimensional tensor following the same approach as \cite{gehrig2023recurrentvisiontransformersobject} and passes them as input to a recurrent vision transformer network \cite{gehrig2023recurrentvisiontransformersobject} pre-trained on event data for object detection. The activations from the first three convolution layers from both input event tensors are compared to yield a quality score that indicates how closely the two event streams match. If performed between simulated and real event streams, we show that this distance is strongly correlated with the simulation gap. The key contributions can be summarized as follows:
\begin{enumerate}    
    \item Propose  \textit{$EQS$}, which, to the best of our knowledge, is the first quantitative and fully differentiable metric that works directly on raw event streams and computes a distance measure.
    \item Show that the $EQS$ metric correlates strongly with how well a model trained on simulated data generalizes to real-world data.        
    \item Demonstrate the importance of bringing simulated events closer to real data, both qualitatively as well as quantitatively in terms of model performance.
\end{enumerate}

% Paper structure:
The rest of the paper is structured as follows. \cref{sec_related:sims} provides an overview of existing event camera simulators and the methodology adopted by each for generating event streams. \cref{sec_related:deepfeats} discusses prior works utilizing deep features for analyzing image data. \cref{sec_motivation} highlights the significance of the simulation gap by examining the drop in performance when testing on real data after training with simulated data. The theoretical details regarding how $EQS$ is computed are explained in \cref{sec_eqs}, and the \cref{sec_experiments} presents empirical results and subsequent analysis for the DSEC \cite{gehrig2021dsecstereoeventcamera} dataset.

%% file: sec/2_related.tex
\begin{figure*}[ht!]
  \centering
    % \fbox{\rule{0pt}{2in} \rule{.9\linewidth}{0pt}}
    \includegraphics[width=0.96\linewidth]{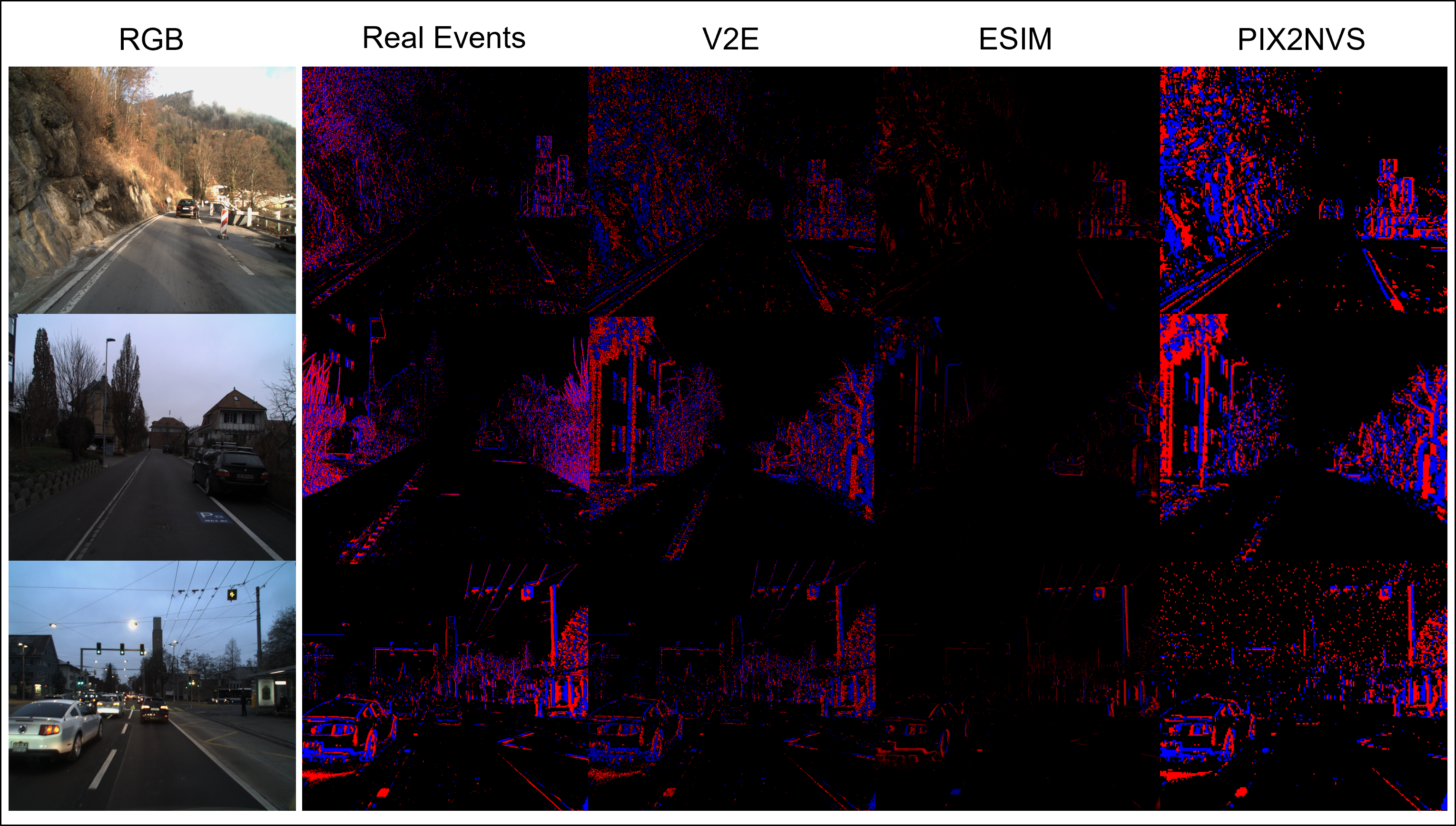}
    \caption{Simulated events from the V2E \cite{hu2021v2evideoframesrealistic}, ESIM  \cite{Rebecq2018}, and PIX2NVS \cite{pix2nvs} event camera simulators and real event camera data visualized as frames where red pixels indicate negative polarity and blue pixels indicate positive polarity.}
    \label{fig:qual_vis}  
\end{figure*}

\section{Related Work}
\label{sec_related}
Initial work on event cameras generally utilized the DVS and DAVIS cameras, followed by more sophisticated sensors developed by Prophesee \cite{prophesee_cams}, iniVation \cite{inivation_cams}, and Lucid Vision Labs \cite{triton_cams}. Early research in event vision was dominated primarily by algorithms to create feature representations of event data. Representations such as time surfaces \cite{Lagorce2017HOTSAH} and HATS \cite{sironi2018hatshistogramsaveragedtime} underscored much of the initial progress in creating machine learning models with event data. As these sensors became more widely available, real-world datasets such as DSEC \cite{gehrig2021dsecstereoeventcamera}, MVSEC \cite{MVSEC} and 1Mpx \cite{perot2020learningdetectobjects1px}, along with simulated datasets like SEVD \cite{aliminati2024sevd} and eTraM \cite{verma2024etram} facilitated training and fine-tuning of larger models on event data.

% Simulators
\subsection{Event Camera Simulators}
\label{sec_related:sims}

% Please add the following required packages to your document preamble:
% \usepackage{graphicx}
\begin{table*}[ht]
\centering
\resizebox{\textwidth}{!}{%
\begin{tabular}{c|c|c|l}
\toprule \toprule
\textbf{Simulator} &
  \textbf{\begin{tabular}[c]{@{}c@{}}Open\\ Source\end{tabular}} &
  \textbf{\begin{tabular}[c]{@{}c@{}}Timestamp\\ Sampling Method\end{tabular}} &
  \multicolumn{1}{c}{\textbf{Methodology}} \\ \midrule \midrule
ESIM \cite{Rebecq2018} &
  \cmark &
  Adaptive &
  Renders synthetic images and uses adaptive sampling for event timing based on brightness changes. Models contrast threshold noise. \\ 
Vid2E \cite{gehrig2020videoeventsrecyclingvideo} &
   \cmark &
  \begin{tabular}[c]{@{}c@{}}Adaptive \\ (via ESIM)\end{tabular} &
  Extends ESIM by using SuperSloMo for temporal interpolation. \\ \midrule
V2E \cite{hu2021v2evideoframesrealistic} &
   \cmark &
  Frame &
  Incorporates bandwidth limitations, leak events, and Poisson shot noise for simulation. \\ 
CARLA \cite{dosovitskiy2017carlaopenurbandriving} &
   \cmark &
  Frame &
  Incorporates uniform noise to pixel differences between consecutive image frames from the rendered $3D$ scene. \\ 
Prophesee Video to Events \cite{prophesee_cams} &
  \xmark &
  Frame &
  Generates event streams using pixel-level interpolation.\\
PIX2NVS \cite{pix2nvs} &
   \cmark &
  Frame &
  Simulates events based on log-intensity or contrast-enhanced differences between successive frames. \\ \midrule
DVS-Voltmeter \cite{Lin2022} &
   \cmark &
  Circuit-based &
  Models voltage noise, photon noise, and temperature noise based on the circuitry of event cameras.\\ 

ADV2E \cite{jiang2024adv2ebridginggapanalogue} &
  \xmark &
  Stochastic &
  Simulates DVS circuit behaviors using continuity sampling and analogue low-pass filtering.
\\ \midrule
PECS \cite{han2024physical} &
  \xmark &
  N/A &
  Works on $3D$ scenes in Blender. Models lens behavior, multispectral rendering, and photocurrent generation using ray tracing. \\ 
ICNS \cite{ICNS} &
   \cmark & Gaussian Noise &
  Performs latency estimation using a delayed filter and light-dependent time constant.\\ 
EventGAN \cite{Zhu2021} &
  \cmark &
  N/A &  
  Learning-based simulator that utilizes Generative Adversarial Networks (GANs) to produce event frames. \\ 
Gazebo DVS Plugin \cite{gazebodvs} &
   \cmark & Frame
   & Thresholds difference between consecutive frames of a video stream captured from sensors in the Gazebo $3D$ simulation.
   \\ \bottomrule \bottomrule
\end{tabular}%
}
\caption{Summary of existing event camera simulators.}
\label{table:simulators}
\end{table*}

Existing event camera simulators generally employ one of the following strategies: $(1)$ Render entire $3D$ scenes in engines such as Blender, Unreal Engine \cite{unrealengine} or CARLA \cite{dosovitskiy2017carlaopenurbandriving}, place a virtual event camera in the scene and sample events \cite{Rebecq2018} \cite{ICNS} or frames \cite{amini2021vista20opendatadriven} at frequencies similar to those of real event cameras to generate a set of events or $(2)$ Use video or image sequences as input and generate event streams by thresholding log intensity differences \cite{hu2021v2evideoframesrealistic} \cite{pix2nvs}. Recently, \cite{Zhu2021} proposed an end-to-end event generation model, and \cite{ICNS, cond_norm_flow, han2024physical} developed physics-aware simulators by parameterizing the event generation process.

The ESIM \cite{Rebecq2018} simulator takes a set of grayscale images as input and performs linear interpolation on the log-intensity values at each pixel to reconstruct a piecewise linear approximation of the underlying signal and adaptively samples this interpolated signal to simulate events. In contrast to previous works such as \cite{Mueggler2017}, which performed uniform sampling on this interpolated signal to generate events, ESIM samples frames adaptively based on the estimated motion field map or optical flow between sampled images. Effectively, it samples events at a higher frequency from sequences with fast motion and lower frequency when the image signal varies slowly. In the absence of ground truth camera trajectories, the adaptive sampling scheme based on pixel displacement detailed in \cite{Rebecq2018} may be used, wherein the next sampling time $t_{k+1}$ from $t_k$ is chosen as:
\[
t_{k+1} = t_k + \lambda_v \left| \mathcal{V}(x; t_k) \right|_m^{-1}
\]
where $\left|\mathcal{V} \right| = \max_{x\in \Omega} \left| \mathcal{V}(x; t_k) \right|$ is the maximum magnitude of the motion field at time $t_k$ and $\lambda_v=0.5$ in the setup detailed in \cite{Rebecq2018}.

Instead of utilizing the motion between frames, the V2E \cite{hu2021v2evideoframesrealistic} simulator seeks to realistically model the photon noise inherent in event cameras using a DVS pixel model that includes temporal noise, leak events, and a finite intensity-dependent bandwidth. The V2E toolbox converts RGB video frames to events by $(1)$. Converting color to luma and $(2)$. Optionally interpolating the luma frames to increase the temporal resolution of the video, $(3)$. Performing a linear to logarithmic mapping by using a linear mapping for luma values $L < 20 $ and a logarithmic scale for larger values. This arises from the observation that DVS pixel bandwidth is proportional to intensity for low photocurrents. This is followed by $(4)$. The event generation model, where it is assumed that each pixel has a memorized brightness value $L_{m}$, and the number of events $N_e =\lfloor \frac{L_{lp}-L_{m}}{\theta} \rfloor$ where $\theta$ varies with a Gaussian distribution with a value drawn from $0.3 + \mathcal{N}(0,\sigma_\theta)$ with $\sigma_\theta = 0.03$. The V2E simulator also models \textit{‘hot pixels’}, which fire events even in the absence of intensity change and spontaneous ON events called \textit{‘leak noise events’} by continuously decreasing the value of $L_m$ by a random amount at each pixel location.

Another relatively early approach, the pixel-to-NVS simulator (PIX2NVS) \cite{pix2nvs} generates event tuples $E_e= \langle x_e, y_e, t_e,P_e \rangle$ from a sequence of pixel-domain video frames $F_0, F_1, ...,F_N$. The RGB pixel values are first converted to luminance values using $y_{i,j}= 0.299R_{i,j} + 0.587 G_{i,j} + 0.114B_{i,j} $, which are then converted to log-intensity using a log-linear mapping:
\[
l_{i,j} = \begin{cases} 
y_{i,j}, & y_{i,j} \leq T_{\text{log}} \\ 
\ln(y_{i,j}), & y_{i,j} > T_{\text{log}}
\end{cases}
\]
where $T_{log}$ threshold controls the switch between linear and log mapping. For lin-log intensity, it is set to a value approximately $10\%$ of the maximum value. For event generation, the difference between luminance values and the average of the weighted neighborhood of luminance values in the previous frame is calculated, and the polarity is determined by the sign of this difference:
\[
d_{i,j}= l_{i,j} - \frac{
\sum_{p=0}^{1} l_{i+2p-1,j}[n-1] + \sum_{p=0}^{1} l_{i,j+2p-1}[n-1]
}{4}
\]
\[
P_e = \begin{cases}
    ON, & \operatorname{sgn}(d_{i,j}) = 1\\
    OFF, & \operatorname{sgn}(d_{i,j}) = -1\\
\end{cases}
\]
The corresponding event is generated if $|d_{i,j}|$ exceeds a fixed threshold and is added to the set of events between the two consecutive frames considered. The event data produced by each of the simulators is qualitatively compared in \cref{fig:qual_vis}.

\subsection{Deep Feature Spaces}
\label{sec_related:deepfeats}
In contrast to traditional frame-based data, event data are characteristically sparse in the spatial domain and much denser in the temporal domain. Thus, effective feature representations must capture both the spatial and temporal aspects of the event stream. The recently proposed recurrent vision transformer (RVT) \cite{gehrig2023recurrentvisiontransformersobject} mixes spatial and temporal features, utilizing transformer layers for spatial feature extraction and recurrent neural networks for temporal feature extraction. Although only activations from the initial convolution layers have been taken for quality evaluation, it should be noted that the LSTM hidden states may also potentially be utilized for a stronger emphasis on temporal features.

% Deep features
It has been shown that feature space division \cite{Kansizoglou_2022} occurs when deep convolutional neural networks are trained on classification tasks, which leads to the activations diverging for each target class after training. It is reasonable to assume that the output of the penultimate layer would refer to a vector that captures the properties of the input in abstract dimensions, such as perceptual quality \cite{lpips}, aesthetics \cite{jayasumana2024rethinkingfidbetterevaluation}, or faithfulness to the text prompt \cite{chen2025contrastivelocalizedlanguageimagepretraining}. More recently, this approach has been utilized to estimate the distances between the distributions of real images and those generated by an algorithm. The classical approach for this is the Frechet Inception Distance \cite{FrechetDistance}, which estimates the distance between a distribution of Inception-v3 features of real and generated images.

%% file: sec/3_EQS.tex
\section{The Motivation}
\label{sec_motivation}

A qualitative visualization of real and simulated events is shown in \cref{fig:qual_vis}. The visualization parameters were kept constant for each event stream to aid a fair comparison, despite each simulator generating a different number of events for the same input image sequence. Qualitatively, it can be seen that the nature of noise in the real event camera data is significantly different from that of the artificial noise added by each simulator. It should be noted that the ESIM output appears darker due to a lower number of events generated at the same pixel position compared to the V2E and PIX2NVS simulators. 

By keeping the number of events plotted constant, the objective is to bring out the differences in how the events are distributed in the simulated event streams compared to that of the Prophesee Gen $3.1$ camera. We hypothesize that these differences in the noise models between real and simulated events cause deep learning models trained on simulated event data to overfit to the noise, which results in sub-optimal features being extracted when tested on real data. This is substantiated by establishing a baseline performance gap by training an RVT-small model on synthetic data produced by each simulator and testing on the actual DSEC test set. These results are summarized in \cref{exp_dsec}. Notably, the higher error on the simulated test set for the model trained on PIX2NVS data indicates that the simulated data fails to represent objects of interest in the synthetic event stream. From the mAP scores reported in \cref{exp_dsec}, it can be seen that the accuracy drops by around $50\%$ when transitioning from simulated to real data for V2E and ESIM, and a much larger gap of about $69\%$ in the case of PIX2NVS. Intuitively, a good event quality measure should be able to detect this and assign a lower score to the event stream generated by this simulator. It is demonstrated in \cref{sec_experiments} that the corresponding $EQS$ values agree with this notion.

\begin{figure*}[ht]
  \centering
    % \fbox{\rule{0pt}{2in} \rule{.9\linewidth}{0pt}}
    % \includegraphics[scale=0.55]{sec/imgs/Arch_v1.png}
    \includegraphics[width=\linewidth]{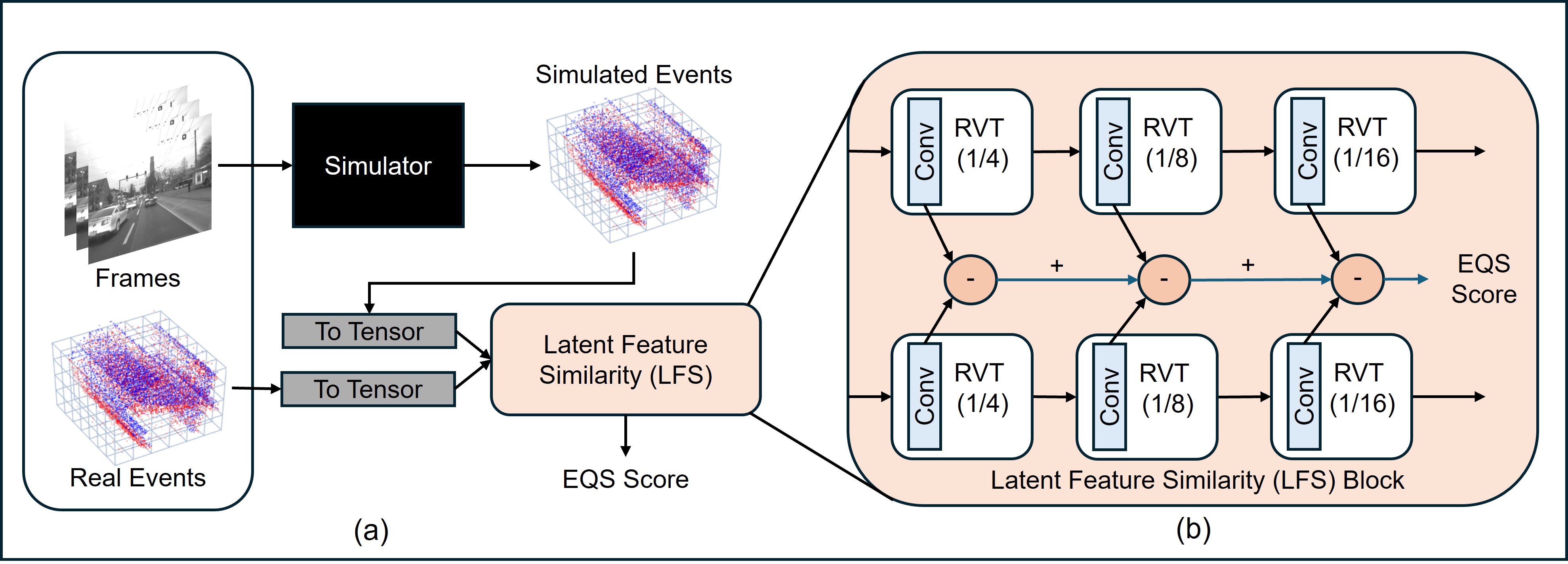 }
    \caption{(a). Dataset with frames and their corresponding synchronized event streams. Simulated events are generated using frames as input, and both sets of events are composed into tensors as discussed in \cref{event_proc}. (b). For each scale in the RVT model, cosine distances are calculated between the feature maps from corresponding convolution layers and aggregated to obtain the final score.}
    \label{fig:Overall_pipeline}  
\end{figure*}

\section{Event Quality Score (EQS)}
\label{sec_eqs}
To work around the severe lack of high-quality event camera data for common tasks in computer vision, such as pose estimation and object detection, several simulators have been proposed, such as V2E \cite{hu2021v2evideoframesrealistic}, ESIM \cite{Rebecq2018}, and end-to-end models like EventGAN \cite{Zhu2021} have been proposed. The quality of the data generated by such methods is usually demonstrated indirectly by showcasing the accuracy of the model trained on simulated data on some downstream task. However, this evaluation is also hindered by the lack of labeled real event camera data. To the best of our knowledge, no numerical metric exists that directly computes a similarity score between two sets of raw events. This brings us to the question: \textit{How to meaningfully measure the distance between two event streams?}

The evaluation metric proposed works directly on the generated event streams. The benefits of directly comparing event streams are twofold: (1) Both spatial and temporal information can be considered in the evaluation as we do not convert events to an image-like representation, and (2) the metric becomes independent of the downstream task, such as object detection or action recognition. The features extracted by a recurrent vision transformer network are used to perform this computation. The mean cosine similarity between the extracted features from each input event stream is taken as the similarity score.

\label{event_proc} Let $E = \{e_{t_1}, e_{t_2}, ..., e_{t_n}\} $ represent a stream of events where each event $e_{t_k} = (x_k, y_k, t_k, p_k)$ is a tuple with polarity $p_k \in \{1,-1\}$ and $|t_n - t_1| \simeq \Delta T$. An event tensor representation is created from event streams in a manner similar to \cite{gehrig2023recurrentvisiontransformersobject}. $T$ temporal bins are created and the positive and negative polarity events that occur within a fixed time interval $\delta t$ are accumulated separately to create a tensor $X_1$ with shape $(2T, H, W)$ where $H$ and $W$ are the height and width of the event camera sensor. This tensor representation is compatible with $2D$ convolutions and can be taken as input to any model with convolution operations, such as Yolo or recurrent vision transformer. The overall architecture for our framework is schematically shown in \cref{fig:Overall_pipeline}.

\subsection{Latent Feature Similarity (LFS) Block}
\label{method_score}
First, event tensors are created for real and simulated event data using the method described in \cref{event_proc} and each tensor is passed to a pre-trained, deep convolutional neural network. Our approach is built on the hypothesis that the sim-to-real gap for event-camera data can be captured in the representational space of the network's activations. The feature maps generated by the convolution layers of the model are extracted and the cosine distance across spatial and temporal features of the network are computed. Although the proposed method can be adapted for all the activations across a given model, we limit our analysis to convolution layers for interpretability. 

Both real and simulated data are passed through a pre-trained recurrent vision transformer network, leveraging both its spatial and temporal feature extraction capabilities. Specifically, the LFS block takes these two sets of activations to compute a similarity score. We extract the feature maps produced by each convolution block and create a $1D$ tensor across depth channels by averaging over patches in the spatial domain. For the RVT model, the choice of LFS patch size for each scale is discussed in \cref{sec_experiments}. Let $Z_i$ be the output of the self-attention block for the $i^{th}$ RVT block in the original formulation from \cite{gehrig2023recurrentvisiontransformersobject}. As shown in \cref{fig:Overall_pipeline} the RVT block at each scale $i$ computes:
\[ (h_i, c_i) = ConvLSTM(Z_i, h_{i-1}, c_{i-1})  \]
\[ o_i = \sigma( Conv(Z_i) + Conv(h_i) )\]
\[ X_{i+1} = h_i + Conv(\sigma(Conv(LN(h_t))))   \]
where $h_i$ and $c_i$ represent the hidden state and cell states respectively, $LN$ denotes layer normalization and $o_i$ is the activation output from the $i^{th}$ scale of the RVT block. 
Let $o_{i_{e1}}$ and $o_{i_{e2}}$ represent the activations from the $i_{th}$ scale of the RVT block for two event streams $e1$ and $e2$ respectively. For each scale, a fixed patch $p_i$ is taken as described in \cref{sec_experiments}. For each non-overlapping patch, the following vector is created
\[ V_{ei} = \left\{ \frac{1}{||p_i||} \sum_{p_i}o_{i_{ei}[p_i]} \right\} \]
such that $||V_{ei}|| $ is the number of channels in $o_{i_{e1}}$. Thus, $ \mathbf{V_{ei}}$ becomes a $1D$ vector for each patch. Taking the vectors $V_{e1}$ and $V_{e2}$, cosine similarity is calculated as usual and subtracted from $1$ to produce a distance metric for each patch.
\[ CS_{p_i} = 1- \frac{ \mathbf{V_{e1}}\cdot \mathbf{V_{e2}} }  {||\mathbf{V_{e1}}||\|\mathbf{V_{e2}}|} \]
The $CS_{p_i}$ scores are averaged across the spatial dimension to obtain a quality score that quantifies the divergence between the two sets of events in latent feature space. It is worth noting that any bounded distance metric can be used for this purpose, but a comprehensive analysis of which distance measure best captures the simulation gap is beyond the scope of this paper. 

%% file: sec/4_Experiments.tex
\section{Experiments}
\label{sec_experiments}

\begin{figure*}[ht!]
  \centering
    \includegraphics[width=\linewidth]{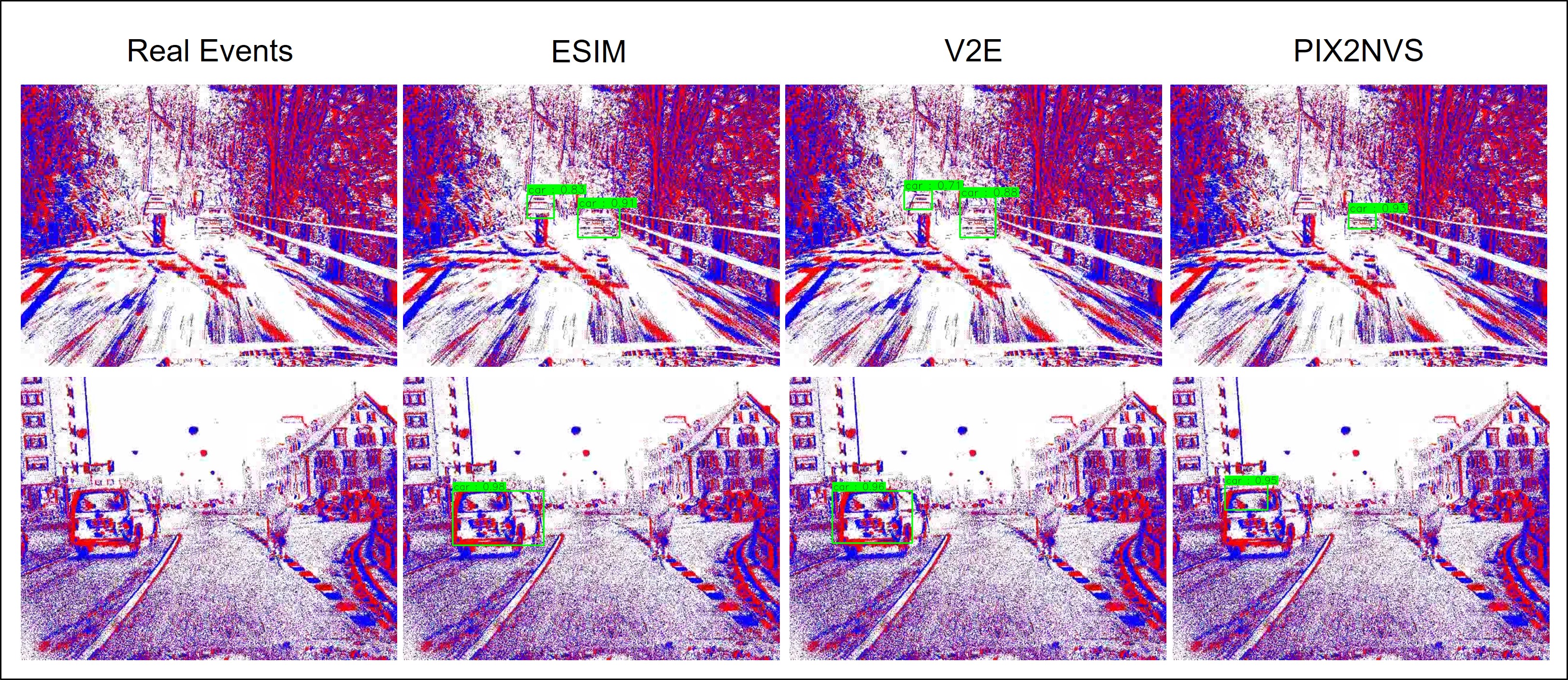 }
    \caption{Detection output samples for RVT-small model on real event streams after training on simulated datasets generated using the ESIM, V2E, and PIX2NVS simulators.}
    \label{fig:qual_detection}  
\end{figure*}

% \subsection{Generating events from frames}
For a set of synchronized RGB images and event streams, the DSEC \cite{gehrig2021dsecstereoeventcamera} dataset was selected. DSEC is a stereo dataset consisting of a pair of Prophesee Gen3.1 event cameras with a resolution of $640 \times 480$ px synchronized with RGB cameras with $1440 \times 1080$ px resolution and a frame rate of $20$ Hz. The RGB frames are converted to grayscale and taken as input to each simulator for generating event streams. For the events generated by the following simulators: ESIM \cite{Rebecq2018}, V2E \cite{hu2021v2evideoframesrealistic}, and PIX2NVS \cite{pix2nvs}, we compute the $EQS$ between the simulated and real event streams.

The DSEC dataset provides RGB frames along with the corresponding synchronized event data. This is important for evaluating event simulation accuracy as simulated event datasets are generated with ESIM \cite{Rebecq2018}, V2E \cite{hu2021v2evideoframesrealistic} and the Pix2NVS \cite{pix2nvs} simulators, and the $EQS$ is computed between each set of simulated event streams and real event camera data. Notably, DSEC is also a stereo dataset, but for the experiments covered here, the data from one set of sensors is sufficient. To evaluate the fidelity of these simulated datasets, the quality score for the entire DSEC dataset and the same for the first $300$ frames of five representative sequences is shown in order to highlight the inter-sequence variation. The results are summarized in \cref{sim_EQS}. As shown in \cref{fig:Overall_pipeline}, we utilize the activations from the conv layers in the first three RVT blocks that downscale the spatial dimension by factors of  $1/4$, $1/8$, and $1/16$, respectively. For each feature map, the spatial dimension is divided into equal chunks of $(3\times 3)$ patches with zero padding if necessary and the cosine similarity score is averaged across all patches. We report the $EQS$ for each simulator over each sequence, where the mean across all sequences can be considered to be the representative score for each simulator.

% This systematically analyzes the performance of each simulator, highlighting differences in their abilities to replicate the temporal and spatial characteristics of real-world event streams.

\begin{table}[]
\centering
\begin{tabular}{c|ccc}
\toprule \midrule
\multirow{2}{*}{\textbf{DSEC Sequence}} & \multicolumn{3}{c}{\textbf{EQS}}                                     \\ \cline{2-4} 
                                        & \multicolumn{1}{c|}{\textbf{V2E}}   & \multicolumn{1}{c|}{\textbf{ESIM}}  & \textbf{Pix2NVS} \\ \midrule \midrule
interlaken\_00\_c                       & \multicolumn{1}{c|}{0.66}           & \multicolumn{1}{c|}{0.72}           & 0.44             \\ \hline
zurich\_city\_01\_e                     & \multicolumn{1}{c|}{0.83}           & \multicolumn{1}{c|}{0.88}           & 0.52             \\ \hline
zurich\_city\_04\_f                     & \multicolumn{1}{c|}{0.74}           & \multicolumn{1}{c|}{0.79}           & 0.61             \\ \hline
zurich\_city\_14\_b                     & \multicolumn{1}{c|}{0.81}           & \multicolumn{1}{c|}{0.91}           & 0.74             \\ \hline
thun\_01\_a                             & \multicolumn{1}{c|}{0.75}           & \multicolumn{1}{c|}{0.84}           & 0.69             \\ \hline
\textbf{Average}                        & \multicolumn{1}{c|}{0.758} & \multicolumn{1}{c|}{\textbf{0.828}} & 0.599  \\ \hline
\textbf{DSEC Average}                   & \multicolumn{1}{c|}{0.773} & \multicolumn{1}{c|}{\textbf{0.866}} & 0.501 \\
\bottomrule
\end{tabular}
\caption{The $EQS$ for each simulator over the first $300$ frames of $5$ randomly chosen sequences and across the entire DSEC \cite{gehrig2021dsecstereoeventcamera} dataset}
\label{sim_EQS}
\end{table}

\begin{table}[]
\centering
\begin{tabular}{c|cc}
\toprule \midrule
\multirow{2}{*}{\textbf{Simulator}} & \multicolumn{2}{c}{\textbf{DSEC Test Accuracy (mAP)}}  \\ \cline{2-3} 
                                    & \multicolumn{1}{c|}{\textbf{Simulated}} & \textbf{Real} \\ \midrule
                               \midrule
V2E                                 & \multicolumn{1}{c|}{37.1}               & 14.3          \\ \hline
ESIM                                & \multicolumn{1}{c|}{\textbf{34.4}}               & \textbf{18.6}          \\ \hline
PIX2NVS                             & \multicolumn{1}{c|}{17.1}               & 05.4          \\ \midrule \bottomrule
\end{tabular}
\caption{Comparing performance of the RVT-small model on the original DSEC test set after training on event data simulated  from frames}
\label{exp_dsec}
\end{table}

\subsection{Evaluating Generalization Performance}
All sequences from the DSEC dataset are run through each simulator and corresponding train, test, and validation sets are generated. The RVT-small \cite{gehrig2023recurrentvisiontransformersobject} model is trained from scratch on this data and tested on the original DSEC test set. \cref{exp_dsec} summarizes the performance of each simulator on the simulated and real test sets from the DSEC dataset. Qualitatively observing the data samples generated by each simulator in \cref{fig:qual_vis} as well as the $EQS$  generated by our quality metric, we can see that the ESIM simulator achieves significantly superior performance on the DSEC test set alongside a higher $EQS$. This shows that the ESIM simulator is better at mimicking the noise model found in the data gathered by the Prophesee sensor compared to V2E and PIX2NVS, as predicted by the higher $EQS$ in \cref{sim_EQS}.

%% file: sec/5_Conclusion.tex
\section{Conclusion and Future Work}
Although a wide range of simulators have been proposed that either generate event-based data from frames or render a virtual $3D$ environment, current simulators are restricted by the simplicity of their event-generation algorithms that largely ignore the nuances of actual event generation. To the best of our knowledge, $EQS$ is the first fully differentiable metric that directly measures how closely a simulated event camera stream mimics real data. Effectively, the $EQS$ quantifies the distance between two raw event camera streams, analogous to the KL-divergence loss commonly used to train generative models. We train a detection model on simulated event data and evaluate it on a real-world driving dataset to demonstrate that $EQS$ is tightly correlated to the model's generalization capabilities to real-world data. Therefore, we anticipate that this can potentially be used as a loss function to optimize machine learning models that take image frames as input and generate corresponding event streams that are close in distribution to real event data. Thus, the $EQS$ can be used to guide the development of the next generation of event-camera simulators, which has the potential to mitigate the difficulties caused by the lack of large labeled event-camera datasets for specific applications in computer vision.

\section{Acknowledgment}
This research is sponsored by NSF Pathways to Enable
Open-Source Ecosystems grant (\#2303748) and the
Partnerships for Innovation grant (\#2329780). We thank data collection support from the Institute of Automated Mobility and computing support from ASU research computing.

%% file: main.bbl
\begin{thebibliography}{50}
\providecommand{\natexlab}[1]{#1}
\providecommand{\url}[1]{\texttt{#1}}
\expandafter\ifx\csname urlstyle\endcsname\relax
  \providecommand{\doi}[1]{doi: #1}\else
  \providecommand{\doi}{doi: \begingroup \urlstyle{rm}\Url}\fi

\bibitem[Aliminati et~al.(2024)Aliminati, Chakravarthi, Verma, Vaghela, Wei, Zhou, and Yang]{aliminati2024sevd}
Manideep~Reddy Aliminati, Bharatesh Chakravarthi, Aayush~Atul Verma, Arpitsinh Vaghela, Hua Wei, Xuesong Zhou, and Yezhou Yang.
\newblock Sevd: Synthetic event-based vision dataset for ego and fixed traffic perception.
\newblock \emph{arXiv preprint arXiv:2404.10540}, 2024.

\bibitem[Amini et~al.(2021)Amini, Wang, Gilitschenski, Schwarting, Liu, Han, Karaman, and Rus]{amini2021vista20opendatadriven}
Alexander Amini, Tsun-Hsuan Wang, Igor Gilitschenski, Wilko Schwarting, Zhijian Liu, Song Han, Sertac Karaman, and Daniela Rus.
\newblock Vista 2.0: An open, data-driven simulator for multimodal sensing and policy learning for autonomous vehicles, 2021.

\bibitem[Bi and Andreopoulos(2017)]{pix2nvs}
Yin Bi and Yiannis Andreopoulos.
\newblock Pix2nvs: Parameterized conversion of pixel-domain video frames to neuromorphic vision streams.
\newblock In \emph{2017 IEEE International Conference on Image Processing (ICIP)}, pages 1990--1994, 2017.

\bibitem[Chakravarthi et~al.(2023)Chakravarthi, Manoj~Kumar, and Pavan~Kumar]{chakravarthi2023event}
Bharatesh Chakravarthi, M Manoj~Kumar, and BN Pavan~Kumar.
\newblock Event-based sensing for improved traffic detection and tracking in intelligent transport systems toward sustainable mobility.
\newblock In \emph{International Conference on Interdisciplinary Approaches in Civil Engineering for Sustainable Development}, pages 83--95. Springer, 2023.

\bibitem[Chakravarthi et~al.(2024)Chakravarthi, Verma, Daniilidis, Fermuller, and Yang]{chakravarthi2024recent}
Bharatesh Chakravarthi, Aayush~Atul Verma, Kostas Daniilidis, Cornelia Fermuller, and Yezhou Yang.
\newblock Recent event camera innovations: A survey.
\newblock \emph{arXiv preprint arXiv:2408.13627}, 2024.

\bibitem[Chen et~al.(2025)Chen, Lai, Zhang, Wang, Eichner, You, Cao, Zhang, Yang, and Gan]{chen2025contrastivelocalizedlanguageimagepretraining}
Hong-You Chen, Zhengfeng Lai, Haotian Zhang, Xinze Wang, Marcin Eichner, Keen You, Meng Cao, Bowen Zhang, Yinfei Yang, and Zhe Gan.
\newblock Contrastive localized language-image pre-training, 2025.

\bibitem[Dosovitskiy et~al.(2017)Dosovitskiy, Ros, Codevilla, Lopez, and Koltun]{dosovitskiy2017carlaopenurbandriving}
Alexey Dosovitskiy, German Ros, Felipe Codevilla, Antonio Lopez, and Vladlen Koltun.
\newblock Carla: An open urban driving simulator, 2017.

\bibitem[{Epic Games}()]{unrealengine}
{Epic Games}.
\newblock Unreal engine.

\bibitem[Gallego et~al.(2022)Gallego, Delbrück, Orchard, Bartolozzi, Taba, Censi, Leutenegger, Davison, Conradt, Daniilidis, and Scaramuzza]{gallego_survey}
Guillermo Gallego, Tobi Delbrück, Garrick Orchard, Chiara Bartolozzi, Brian Taba, Andrea Censi, Stefan Leutenegger, Andrew~J. Davison, Jörg Conradt, Kostas Daniilidis, and Davide Scaramuzza.
\newblock Event-based vision: A survey.
\newblock \emph{IEEE Transactions on Pattern Analysis and Machine Intelligence}, 44\penalty0 (1):\penalty0 154--180, 2022.

\bibitem[Gehrig et~al.(2020)Gehrig, Gehrig, Hidalgo-Carrió, and Scaramuzza]{gehrig2020videoeventsrecyclingvideo}
Daniel Gehrig, Mathias Gehrig, Javier Hidalgo-Carrió, and Davide Scaramuzza.
\newblock Video to events: Recycling video datasets for event cameras, 2020.

\bibitem[Gehrig and Scaramuzza(2023)]{gehrig2023recurrentvisiontransformersobject}
Mathias Gehrig and Davide Scaramuzza.
\newblock Recurrent vision transformers for object detection with event cameras, 2023.

\bibitem[Gehrig et~al.(2021)Gehrig, Aarents, Gehrig, and Scaramuzza]{gehrig2021dsecstereoeventcamera}
Mathias Gehrig, Willem Aarents, Daniel Gehrig, and Davide Scaramuzza.
\newblock Dsec: A stereo event camera dataset for driving scenarios, 2021.

\bibitem[Gu et~al.(2024)Gu, Li, Zhu, Zhang, and Ren]{cond_norm_flow}
Daxin Gu, Jia Li, Lin Zhu, Yu Zhang, and Jimmy~S. Ren.
\newblock Reliable event generation with invertible conditional normalizing flow.
\newblock \emph{IEEE Transactions on Pattern Analysis and Machine Intelligence}, 46\penalty0 (2):\penalty0 927--943, 2024.

\bibitem[Han et~al.(2024)Han, Lyu, Li, Wei, Li, Wei, Chen, and Ji]{han2024physical}
Haiqian Han, Jiacheng Lyu, Jianing Li, Henglu Wei, Cheng Li, Yajing Wei, Shu Chen, and Xiangyang Ji.
\newblock Physical-based event camera simulator.
\newblock In \emph{European Conference on Computer Vision}, pages 19--35. Springer, 2024.

\bibitem[Heusel et~al.(2018)Heusel, Ramsauer, Unterthiner, Nessler, and Hochreiter]{FrechetDistance}
Martin Heusel, Hubert Ramsauer, Thomas Unterthiner, Bernhard Nessler, and Sepp Hochreiter.
\newblock Gans trained by a two time-scale update rule converge to a local nash equilibrium, 2018.

\bibitem[Hu et~al.(2021)Hu, Liu, and Delbruck]{hu2021v2evideoframesrealistic}
Yuhuang Hu, Shih-Chii Liu, and Tobi Delbruck.
\newblock v2e: From video frames to realistic dvs events, 2021.

\bibitem[Inivation()]{inivation_cams}
Inc Inivation.
\newblock Inivation.
\newblock \url{https://inivation.com/solutions/cameras/}.

\bibitem[Jayasumana et~al.(2024)Jayasumana, Ramalingam, Veit, Glasner, Chakrabarti, and Kumar]{jayasumana2024rethinkingfidbetterevaluation}
Sadeep Jayasumana, Srikumar Ramalingam, Andreas Veit, Daniel Glasner, Ayan Chakrabarti, and Sanjiv Kumar.
\newblock Rethinking fid: Towards a better evaluation metric for image generation, 2024.

\bibitem[Jiang et~al.(2024)Jiang, Zhou, and Lin]{jiang2024adv2ebridginggapanalogue}
Xiao Jiang, Fei Zhou, and Jiongzhi Lin.
\newblock Adv2e: Bridging the gap between analogue circuit and discrete frames in the video-to-events simulator, 2024.

\bibitem[Joubert et~al.(2021)Joubert, Marcireau, Ralph, Jolley, van Schaik, and Cohen]{ICNS}
Damien Joubert, Alexandre Marcireau, Nic Ralph, Andrew Jolley, André van Schaik, and Gregory Cohen.
\newblock Event camera simulator improvements via characterized parameters.
\newblock \emph{Frontiers in Neuroscience}, 15:\penalty0 702765, 2021.

\bibitem[Kaiser et~al.(2016)Kaiser, Tieck, Hubschneider, Wolf, Weber, Hoff, Friedrich, Wojtasik, Roennau, Kohlhaas, Dillmann, and Zöllner]{gazebodvs}
J. Kaiser, J.~C.~V. Tieck, C. Hubschneider, P. Wolf, M. Weber, M. Hoff, A. Friedrich, K. Wojtasik, A. Roennau, R. Kohlhaas, R. Dillmann, and J.~M. Zöllner.
\newblock Towards a framework for end-to-end control of a simulated vehicle with spiking neural networks.
\newblock In \emph{2016 IEEE International Conference on Simulation, Modeling, and Programming for Autonomous Robots (SIMPAR)}, pages 127--134, 2016.

\bibitem[Kansizoglou et~al.(2022)Kansizoglou, Bampis, and Gasteratos]{Kansizoglou_2022}
Ioannis Kansizoglou, Loukas Bampis, and Antonios Gasteratos.
\newblock Deep feature space: A geometrical perspective.
\newblock \emph{IEEE Transactions on Pattern Analysis and Machine Intelligence}, 44\penalty0 (10):\penalty0 6823–6838, 2022.

\bibitem[Kim et~al.(2022)Kim, Bae, Park, Zhang, and Kim]{kim2022nimagenetrobustfinegrainedobject}
Junho Kim, Jaehyeok Bae, Gangin Park, Dongsu Zhang, and Young~Min Kim.
\newblock N-imagenet: Towards robust, fine-grained object recognition with event cameras, 2022.

\bibitem[Lagorce et~al.(2017)Lagorce, Orchard, Galluppi, Shi, and Benosman]{Lagorce2017HOTSAH}
Xavier Lagorce, G. Orchard, Francesco Galluppi, Bertram~E. Shi, and Ryad~B. Benosman.
\newblock Hots: A hierarchy of event-based time-surfaces for pattern recognition.
\newblock \emph{IEEE Transactions on Pattern Analysis and Machine Intelligence}, 39:\penalty0 1346--1359, 2017.

\bibitem[Li et~al.(2024)Li, Wang, Yuan, Li, Weng, Peng, Zhang, Xiong, and Sun]{li2024eventassistedlowlightvideoobject}
Hebei Li, Jin Wang, Jiahui Yuan, Yue Li, Wenming Weng, Yansong Peng, Yueyi Zhang, Zhiwei Xiong, and Xiaoyan Sun.
\newblock Event-assisted low-light video object segmentation, 2024.

\bibitem[Liang et~al.(2022)Liang, Cao, Yang, Zhang, and Chen]{event_kitti}
Zichen Liang, Hu Cao, Chu Yang, Zikai Zhang, and Guang Chen.
\newblock Global-local feature aggregation for event-based object detection on eventkitti.
\newblock In \emph{2022 IEEE International Conference on Multisensor Fusion and Integration for Intelligent Systems (MFI)}, pages 1--7, 2022.

\bibitem[Lichtsteiner et~al.(2008)Lichtsteiner, Posch, and Delbruck]{delbruck_davis}
Patrick Lichtsteiner, Christoph Posch, and Tobi Delbruck.
\newblock A 128$\times$ 128 120 db 15 $\mu$s latency asynchronous temporal contrast vision sensor.
\newblock \emph{IEEE Journal of Solid-State Circuits}, 43\penalty0 (2):\penalty0 566--576, 2008.

\bibitem[Lin et~al.(2022)Lin, Ma, Guo, and Wen]{Lin2022}
Songnan Lin, Ye Ma, Zhenhua Guo, and Bihan Wen.
\newblock Dvs-voltmeter: Stochastic process-based event simulator for dynamic vision sensors.
\newblock In \emph{Lecture Notes in Computer Science (including subseries Lecture Notes in Artificial Intelligence and Lecture Notes in Bioinformatics)}, 2022.

\bibitem[Liu et~al.(2023)Liu, Xu, Yang, Yu, and Yu]{Liu_2023}
Bingde Liu, Chang Xu, Wen Yang, Huai Yu, and Lei Yu.
\newblock Motion robust high-speed light-weighted object detection with event camera.
\newblock \emph{IEEE Transactions on Instrumentation and Measurement}, 72:\penalty0 1–13, 2023.

\bibitem[Liu et~al.(2024)Liu, Peng, Zhu, Chang, Zhou, and Yan]{liu2024seeingmotionnighttimeevent}
Haoyue Liu, Shihan Peng, Lin Zhu, Yi Chang, Hanyu Zhou, and Luxin Yan.
\newblock Seeing motion at nighttime with an event camera, 2024.

\bibitem[Lucid~vision labs()]{triton_cams}
Inc Lucid~vision labs.
\newblock Lucid vision labs.
\newblock \url{ https://thinklucid.com/triton2-evs/ }.

\bibitem[Mueggler et~al.(2017)Mueggler, Rebecq, Gallego, Delbruck, and Scaramuzza]{Mueggler2017}
Elias Mueggler, Henri Rebecq, Guillermo Gallego, Tobi Delbruck, and Davide Scaramuzza.
\newblock The event-camera dataset and simulator: Event-based data for pose estimation, visual odometry, and slam.
\newblock \emph{The International Journal of Robotics Research}, 36:\penalty0 142--149, 2017.

\bibitem[Muglikar et~al.(2023)Muglikar, Bauersfeld, Moeys, and Scaramuzza]{Muglikar23CVPR_esfp}
Manasi Muglikar, Leonard Bauersfeld, Diederik Moeys, and Davide Scaramuzza.
\newblock Event-based shape from polarization.
\newblock In \emph{IEEE / CVF Computer Vision and Pattern Recognition Conference (CVPR)}, 2023.

\bibitem[Muglikar et~al.(2024)Muglikar, Somasundaram, Dave, Charbon, Raskar, and Scaramuzza]{muglikar2024eventcamerasmeetspads}
Manasi Muglikar, Siddharth Somasundaram, Akshat Dave, Edoardo Charbon, Ramesh Raskar, and Davide Scaramuzza.
\newblock Event cameras meet spads for high-speed, low-bandwidth imaging, 2024.

\bibitem[Perot et~al.(2020)Perot, de~Tournemire, Nitti, Masci, and Sironi]{perot2020learningdetectobjects1px}
Etienne Perot, Pierre de Tournemire, Davide Nitti, Jonathan Masci, and Amos Sironi.
\newblock Learning to detect objects with a 1 megapixel event camera, 2020.

\bibitem[Pijnacker~Hordijk et~al.(2017)Pijnacker~Hordijk, Scheper, and de~Croon]{Pijnacker_Hordijk_2017}
Bas~J. Pijnacker~Hordijk, Kirk Y.~W. Scheper, and Guido C. H.~E. de Croon.
\newblock Vertical landing for micro air vehicles using event‐based optical flow.
\newblock \emph{Journal of Field Robotics}, 35\penalty0 (1):\penalty0 69–90, 2017.

\bibitem[Prophesee()]{prophesee_cams}
Inc Prophesee.
\newblock Prophesee.
\newblock \url{ https://www.prophesee.ai/event-based-sensors/ }.

\bibitem[Rebecq et~al.(2018)Rebecq, Gehrig, and Scaramuzza]{Rebecq2018}
Henri Rebecq, Daniel Gehrig, and Davide Scaramuzza.
\newblock Esim: an open event camera simulator.
\newblock In \emph{Proceedings of Machine Learning Research}, 2018.

\bibitem[Rebecq et~al.(2019)Rebecq, Ranftl, Koltun, and Scaramuzza]{rebecq2019eventstovideobringingmoderncomputer}
Henri Rebecq, René Ranftl, Vladlen Koltun, and Davide Scaramuzza.
\newblock Events-to-video: Bringing modern computer vision to event cameras, 2019.

\bibitem[Ronneberger et~al.(2015)Ronneberger, Fischer, and Brox]{ronneberger2015unetconvolutionalnetworksbiomedical}
Olaf Ronneberger, Philipp Fischer, and Thomas Brox.
\newblock U-net: Convolutional networks for biomedical image segmentation, 2015.

\bibitem[Simonyan and Zisserman(2015)]{simonyan2015deepconvolutionalnetworkslargescale_vgg}
Karen Simonyan and Andrew Zisserman.
\newblock Very deep convolutional networks for large-scale image recognition, 2015.

\bibitem[Sironi et~al.(2018)Sironi, Brambilla, Bourdis, Lagorce, and Benosman]{sironi2018hatshistogramsaveragedtime}
Amos Sironi, Manuele Brambilla, Nicolas Bourdis, Xavier Lagorce, and Ryad Benosman.
\newblock Hats: Histograms of averaged time surfaces for robust event-based object classification, 2018.

\bibitem[Stoffregen et~al.(2020)Stoffregen, Scheerlinck, Scaramuzza, Drummond, Barnes, Kleeman, and Mahony]{Stoffregen2020}
Timo Stoffregen, Cedric Scheerlinck, Davide Scaramuzza, Tom Drummond, Nick Barnes, Lindsay Kleeman, and Robert Mahony.
\newblock \emph{Reducing the Sim-to-Real Gap for Event Cameras}, pages 534--549.
\newblock 2020.

\bibitem[Verma et~al.(2024)Verma, Chakravarthi, Vaghela, Wei, and Yang]{verma2024etram}
Aayush~Atul Verma, Bharatesh Chakravarthi, Arpitsinh Vaghela, Hua Wei, and Yezhou Yang.
\newblock etram: Event-based traffic monitoring dataset.
\newblock In \emph{Proceedings of the IEEE/CVF Conference on Computer Vision and Pattern Recognition}, pages 22637--22646, 2024.

\bibitem[Vidal et~al.(2018)Vidal, Rebecq, Horstschaefer, and Scaramuzza]{Vidal_2018}
Antoni~Rosinol Vidal, Henri Rebecq, Timo Horstschaefer, and Davide Scaramuzza.
\newblock Ultimate slam? combining events, images, and imu for robust visual slam in hdr and high-speed scenarios.
\newblock \emph{IEEE Robotics and Automation Letters}, 3\penalty0 (2):\penalty0 994–1001, 2018.

\bibitem[Wang et~al.(2004)Wang, Bovik, Sheikh, and Simoncelli]{ssim}
Zhou Wang, A.C. Bovik, H.R. Sheikh, and E.P. Simoncelli.
\newblock Image quality assessment: from error visibility to structural similarity.
\newblock \emph{IEEE Transactions on Image Processing}, 13\penalty0 (4):\penalty0 600--612, 2004.

\bibitem[Zhang et~al.(2018)Zhang, Isola, Efros, Shechtman, and Wang]{lpips}
Richard Zhang, Phillip Isola, Alexei~A. Efros, Eli Shechtman, and Oliver Wang.
\newblock The unreasonable effectiveness of deep features as a perceptual metric.
\newblock In \emph{2018 IEEE/CVF Conference on Computer Vision and Pattern Recognition}, pages 586--595, 2018.

\bibitem[Zhang et~al.(2023)Zhang, Chai, Yu, Majaj, Walsh, Wang, Mahbub, Siegelmann, Kim, and Rahman]{yelan_syn_dataset}
Zhongyang Zhang, Kaidong Chai, Haowen Yu, Ramzi Majaj, Francesca Walsh, Edward Wang, Upal Mahbub, Hava Siegelmann, Donghyun Kim, and Tauhidur Rahman.
\newblock Neuromorphic high-frequency 3d dancing pose estimation in dynamic environment.
\newblock \emph{Neurocomputing}, 547:\penalty0 126388, 2023.

\bibitem[Zhu et~al.(2018)Zhu, Thakur, Ozaslan, Pfrommer, Kumar, and Daniilidis]{MVSEC}
Alex~Zihao Zhu, Dinesh Thakur, Tolga Ozaslan, Bernd Pfrommer, Vijay Kumar, and Kostas Daniilidis.
\newblock The multivehicle stereo event camera dataset: An event camera dataset for 3d perception.
\newblock \emph{IEEE Robotics and Automation Letters}, 3\penalty0 (3):\penalty0 2032–2039, 2018.

\bibitem[Zhu et~al.(2021)Zhu, Wang, Khant, and Daniilidis]{Zhu2021}
Alex~Zihao Zhu, Ziyun Wang, Kaung Khant, and Kostas Daniilidis.
\newblock Eventgan: Leveraging large scale image datasets for event cameras.
\newblock In \emph{2021 IEEE International Conference on Computational Photography, ICCP 2021}, 2021.

\end{thebibliography}
